\documentclass[runningheads]{llncs}
\usepackage[T1]{fontenc}
\usepackage{marvosym}
\usepackage{mathtools}
\usepackage{graphicx}
\usepackage{booktabs}
\usepackage{multirow}
\usepackage{multicol}
\usepackage{amsmath}
\usepackage{amssymb}
\usepackage{colortbl}
\usepackage{pifont}
\usepackage{CJKutf8}
\usepackage{bbm}
\usepackage{bm}
\usepackage{dsfont}
\usepackage{MnSymbol}
\usepackage{wrapfig}
\usepackage[font=tiny]{subfig}
\usepackage{tabularx}
\usepackage{xcolor}

\definecolor{Gray}{gray}{0.93}
\def\@mb@citenamelist{cite,citep,citet,citealp,citealt,citepalias,citetalias}

\begin{document}
\begin{CJK*}{UTF8}{gbsn}

\title{Explaining Black-box Language Models with Knowledge Probing Systems: \\ A Post-hoc Explanation Perspective}
\titlerunning{Explaining Black-box Language Models with Knowledge Probing Systems}

\author{Yunxiao Zhao\inst{1} \and Hao Xu\inst{1} \and Zhiqiang Wang\inst{1,2}\textsuperscript{(\Letter)} \and Xiaoli Li\inst{3} \and \\ Jiye Liang\inst{1,2} \and  Ru Li\inst{1,2}\textsuperscript{(\Letter)} }
\authorrunning{Y. Zhao et al.}

%
\institute{
School of Computer and Information Technology, Shanxi University, China 
\and Key Laboratory of Computational Intelligence and Chinese Information Processing of Ministry of Education, Shanxi University, China \\ 
\and Institute for Infocomm Research, A*Star, Singapore \\
\email{\{yunxiaomr,xuhao0050\}@163.com} \\
\email{\{wangzq,ljy,liru\}@sxu.edu.cn, }
\email{xlli@ntu.edu.sg}
}
\maketitle  
\begin{abstract} %

Pre-trained Language Models (PLMs) are trained on large amounts of unlabeled data, yet they exhibit remarkable reasoning skills. However, the trustworthiness challenges posed by these black-box models have become increasingly evident in recent years. 
To alleviate this problem, this paper proposes a novel {Know}ledge-guided {Prob}ing approach called KnowProb in a post-hoc explanation way, which aims to probe whether black-box PLMs understand implicit knowledge beyond the given text, rather than focusing only on the surface level content of the text. We provide six potential explanations derived from the underlying content of the given text, including three knowledge-based understanding and three association-based reasoning. 
In experiments, we validate that current small-scale (or large-scale) PLMs only learn a single distribution of representation, and still face significant challenges in capturing the hidden knowledge behind a given text. 
Furthermore, we demonstrate that our proposed approach is effective for identifying the limitations of existing black-box models from multiple probing perspectives, which facilitates researchers to promote the study of detecting black-box models in an explainable way. 
\end{abstract}

\section{Introduction}

Pre-trained language models (PLMs) have revolutionized natural language processing and computer vision tasks, in which the success is attributed to the rich representations and the knowledge captured from the structural and non-structural data \cite{de-cao-etal-2021-editing,ye-etal-2022-zerogen}. 
However, knowledge storage and localization are challenging. 
An emerging line of research increasingly views PLMs as soft knowledge bases \cite{petroni-etal-2019-language,sung-etal-2021-language} and focuses on investigating and quantifying the types and amounts of knowledge they encode \cite{roberts-etal-2020-much,petroni-etal-2019-language}, which is known by knowledge probing. It not only can obtain performance gains on downstream tasks but also improve the explainability of black-box models through a model-agnostic perspective.  

Knowledge probing typically involves a black-box PLM and a probing dataset that contains factual knowledge, including syntax, linguistic patterns, and other facts. This dataset is used to estimate the extent of knowledge embedded in the model \cite{youssef-etal-2023-give}. Probing is often performed using methods such as cloze prompts \cite{chen-etal-2022-meta}, questions \cite{Kalo2022KAMELKA}, or entities \cite{lietard-etal-2021-language,dufter-etal-2021-static,shi-etal-2021-descgen} to extract factual knowledge. 
For example, Bouraoui et al. \cite{bouraoui2020inducing,jiang-etal-2020-know} use diversification and mining methods, employing cloze prompts as input to probe relational and factual knowledge, respectively. Wang et al. \cite{wang-etal-2021-generative} adopt a closed-book QA approach to investigate BART \cite{lewis-etal-2020-bart} as a knowledge base to retain training facts. Additionally, Liétard et al. \cite{lietard-etal-2021-language} use fixed entity inputs, such as locations (e.g., countries or cities), to probe geographical knowledge. As a result, probing black-box PLMs has shown some improvements.

\begin{figure}[t]
    \centering
    \includegraphics[scale=0.42]{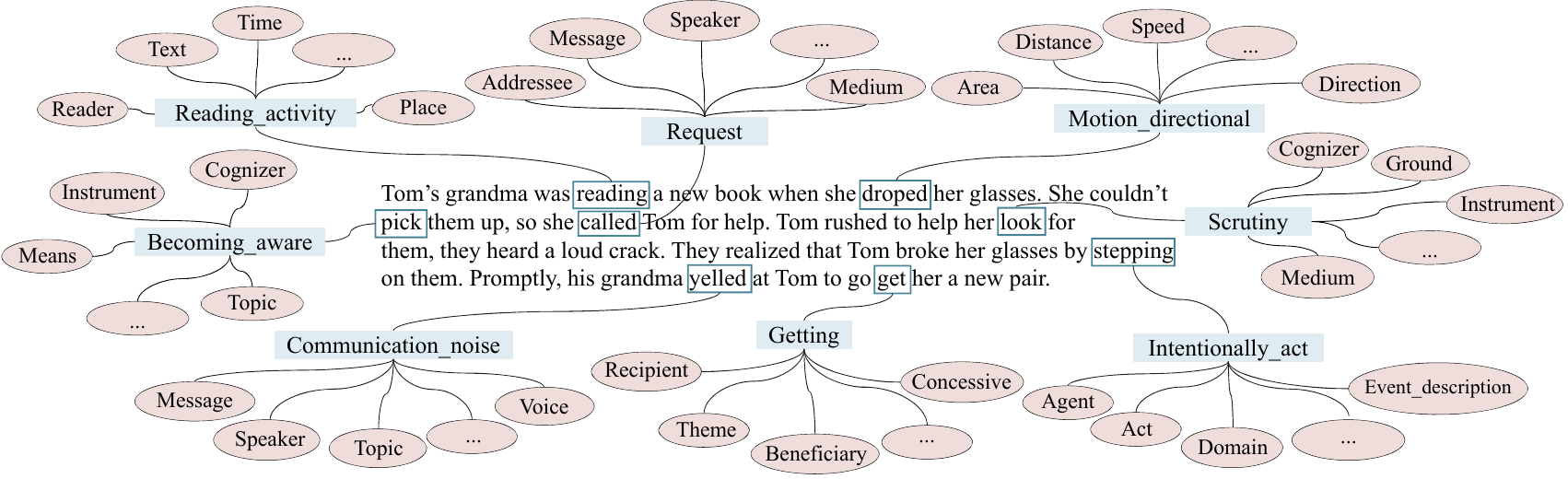} 
	\caption{A natural language text example with hidden knowledge beyond the text itself, where the hidden knowledge is not explicitly present in the given text.} %
    \label{figure1}
\end{figure}

However, an understanding of knowledge is often more profound. In other words, understanding a knowledge concept typically requires a deep comprehension of the underlying knowledge associated with it \cite{FrameCS,li2024comprehensive}. As shown in Figure \ref{figure1}, understanding "reading" often involves constructing a scenario where several pieces of knowledge are present, such as: 1) there is a reader, and 2) the reader receives a message from a text. Existing methods tend to focus more on probing surface-level knowledge within the language text (e.g., reading), while neglecting the deeper, hidden knowledge behind the text (e.g., <reader, read, text>). 

In addition, with the growth of open source datasets and the development of deep learning technologies, especially for small-scale (or large-scale) PLMs like BERTs and GPTs, we have witnessed rapid progress recently, attaining levels of performance akin to or surpassing the average human capabilities across various benchmarks \cite{rajpurkar-etal-2016-squad,gcrc,roberts-etal-2020-much}. 
However, there may be potential data biases and evaluation errors. Lewis et al. \cite{lewis-etal-2021-question} identify risks such as the train-test overlap, and note the absence of mechanisms to control for the knowledge already used to train a PLM before it is tested on closed-book questions \cite{wang-etal-2021-generative}. 

To address these problems, we consider the integration of the FrameNet \cite{baker-etal-1998-berkeley-framenet} and propose \textbf{KnowProb}, a novel \textbf{Know}ledge-based \textbf{Prob}ing with frame modeling to comprehensively estimate a black-box model's ability to understand hidden knowledge behind the given text. In particular, \texttt{KnowProb} is an approach that utilizes frame semantic parser to model and generate entities (or roles) for probing those knowledge beyond the given text such that it can directly improve train-test overlap problem. Following previous research \cite{haviv-etal-2021-bertese,zhang-etal-2022-promptgen}, we construct the input prompts with the help of frame semantic parser-based PLMs. 
To the best of our knowledge, this is the first exploration to investigate factual knowledge based on frame concepts. 
Our contributions can be summarised as follows:

\begin{itemize}
\item[1.] We propose {KnowProb}, a novel {Know}ledge-based {Prob}ing method with frame modeling for probing knowledge beyond the given text, which models how human understand a standard knowledge concept and situation to better estimate models via incorporating semantic frames from FrameNet. 

\item[2.] We validate that current small-scale (or large-scale) PLMs only learn a single distribution of representations, and still face significant challenges in capturing the hidden knowledge behind a given text, which highlights opportunities for improving reasoning across out-of-domain distributions in the future.

\item[3.] Furthermore, we demonstrate the effectiveness of our proposed approach by experiments, which not only aids in identifying the shortcomings of existing PLMs but also does so in a manner that is both explainable (a post-hoc explanation perspective) and rooted in cognitive linguistic principles. 
\end{itemize}

\section{Related Work}
\noindent \textbf{Knowledge Probing. } 
It typically involves a black-box PLM and a probing dataset, with the goal of identifying where knowledge is stored. According to the types of knowledge, there can be categorized into general knowledge \cite{wang-etal-2021-generative,shi-etal-2021-descgen,mallen-etal-2023-trust}, domain-specific knowledge \cite{sung-etal-2021-language,meng-etal-2022-rewire}, and other factual knowledge \cite{margatina-etal-2023-dynamic,lee-etal-2022-plug,meng2022locating}. On the other hand, approaches \cite{chen-etal-2022-meta,Kalo2022KAMELKA,lietard-etal-2021-language} such as utilizing non-optimized inputs and directly probing the capabilities of PLMs have been proposed to investigate black-box models. However, PLMs are sensitive to the inputs \cite{petroni-etal-2019-language}. Thus optimized input methods liking diversification and mining \cite{bouraoui2020inducing}, direct optimization \cite{saeed-papotti-2022-type}, and generation with PLMs \cite{haviv-etal-2021-bertese} are proposed to address this problem. For probed PLMs, they are probed for knowledge using either their original pre-trained parameters \cite{petroni-etal-2019-language,jiang-etal-2020-know}, or after adapting these parameters \cite{roberts-etal-2020-much}. Popular research focuses on models such as BERT \cite{2019bert}, ERNIE \cite{ERNIE3.0}, and GPTs \cite{touvron2023llama}.

\begin{figure}[t]
    \centering
    \includegraphics[scale=0.72]{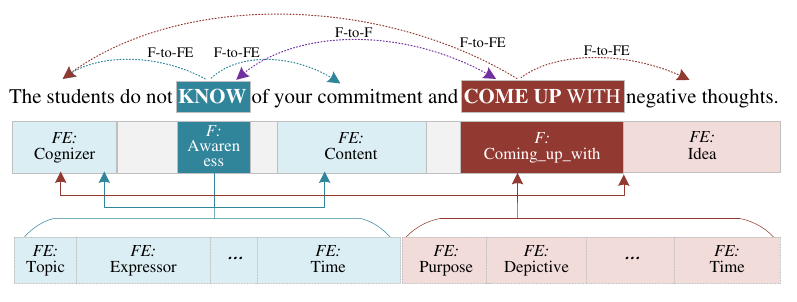}
     \caption{An example with frame representation within a sentence, where there are semantic scenarios described by two frames. In particular, different frames and frame elements can be interconnected through Frame-to-Frame Element (F-to-FE) and Frame Element-to-Frame Element (FE-to-FE) relationships.}  
    \label{figure5}
\end{figure}

\noindent \textbf{FrameNet Database. }
FrameNet is a database with linguistic knowledge, built on the hypothesis that people understand things by performing mental operations on what they already know \cite{baker-etal-1998-berkeley-framenet}. 
This kind of knowledge reflecting the cognitive experience of people is described as structured information packets called \texttt{frames (Fs)} \cite{su-etal-2021-knowledge}.
A frame can represent the knowledge necessary to understand a standard situation \cite{FrameCS}, associated with a set of semantic roles \texttt{(frame elements (FEs))}. \texttt{Lexical units (LUs)} are capable of evoking the scenarios represented by corresponding frames. 
As Figure \ref{figure5} shown, the frame \texttt{Awareness} indicates the scenario \texttt{KNOW}, and it utilizes FEs and LUs to describe, respectively, scenario-related knowledge and the scenario-triggering word. 

\noindent \textbf{Models Explainability.} 
There are two main research lines in the field of black-box explainability. One line focuses on research related to built-in explanations of models, often referred to as \texttt{ante-hoc explanation}, such as self-explanation rationalization. This approach has been extensively studied in areas such as natural language processing \cite{zhao-etal-2024-agr} and computer vision \cite{yuan2020interpreting}. The other line is \texttt{post-hoc explanation}, which typically utilizes methods independent of model training to provide explainable insights for black box models already trained. These methods include probing-based detection \cite{youssef-etal-2023-give}, the use of surrogate models for approximation such as LIME \cite{ribeiro-etal-2016-trust}, and so on.

\section{Preliminaries}
To enable \texttt{KnowProb} to generate knowledge beyond the given text, we first construct a seed dataset\footnote{They are provided in \url{https://github.com/yunxiaomr/KnowProb}.} for supervised fine-tuning. During the construction process, we follow the three steps:

\textbf{Step 1: Collecting Dataset.} 
Our corpus resources come mainly from the Chinese Tree Bank \cite{CTB}, which is a typical corpus, including many texts from newswires, magazines and various broadcast news. 
Firstly, we search and filter about 15k sentences. Then, for preprocessing, we remove those sentences with the sequence length too short for a special frame. 
Next, we annotate the frame-semantic information based on concepts of frame semantics. 

\textbf{Step 2: Data Annotation.} 
Before the annotation work, we first preparean annotation guideline and an annotation system, including annotation standards, etc., in detail. 
During the annotation process, annotators are required to identify the semantic frames in the sentence and frame elements based on those frames. As Figure \ref{figure5} shown, annotators first need to identify the \texttt{Awareness} and \texttt{Come\_up\_with}\footnote{https://framenet.icsi.berkeley.edu/}. Next, for \texttt{Awareness}, they need to annotate their frame elements and mark them using the filling text of the sentence. 

\textbf{Step 3: Quality Assurance}. 
We implement a rigorous annotation and auditing protocol, including three iterative rounds. In instances where annotators encounter uncertainties, we facilitate discussions with linguistic experts, culminating in definitive annotations. The outcomes of this consistent annotation process are then integrated into the resource database. 
In total, we construct a frame base including 10,000 lexical units, 695 frames and 990 FEs in this paper. 
For sentence instances with frame semantic, we annotate 10,000 sentences, involving 629 instances of frame and 673 instances of frame elements. 

\section{Methodology}
\begin{figure*}[t]
    \centering
    \includegraphics[scale=0.12]{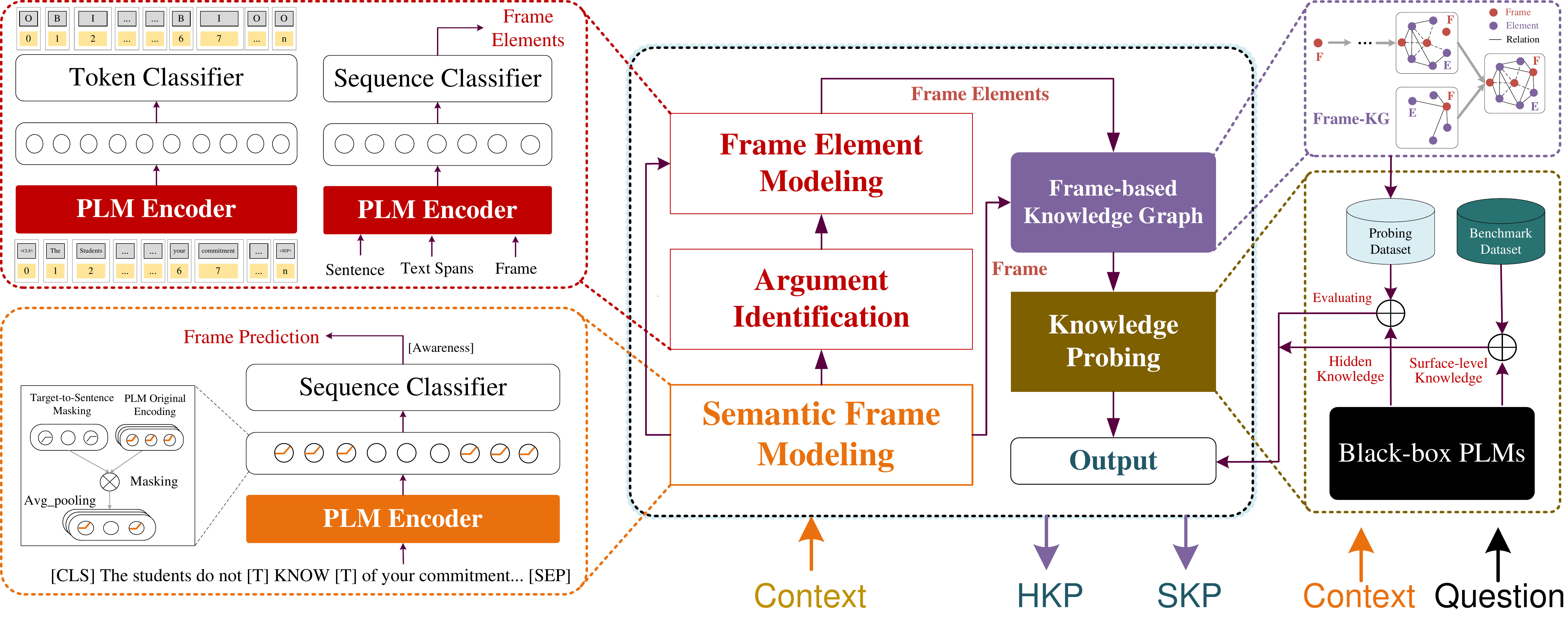} 
     \caption{The overview of KnowProb. Considering a MRC-QA task, it takes the QA context and question as input. The HKP (Hidden Knowledge Performance) and SKP (Surface-level Knowledge Performance) denote the performance of hidden knowledge beyond the text itself and the traditional QA performance.}
    \label{figure2}
\end{figure*}

As Figure \ref{figure2} shows, we present the overall architecture of \texttt{KnowProb} which utilizes semantic frames to model hidden knowledge beyond the text itself, including the process of knowledge generation and knowledge probing. 

\subsection{Frame Semantic Parser}

\subsubsection{Semantic Frame Modeling}
We utilize the frame identification task to perform semantic frame modeling. Frame identification aims to predict a frame evoked by a target word in a sentence \cite{su-etal-2021-knowledge}, which can be defined as a mapping function $ G: f(s,t) \to f_j$,  subject to $f_j \in F$, where $s$ and $t$ denote the sentence and the target, respectively.
We first decompose the context to obtain a set of sentences $S=\{s_0,...,s_{k-1}\}$. Then we use a retrieval-based method to identify target words in the sentence $s_i$ where $i \in [0,k)$, 
where those lexical units retrieved are manually annotated and mapped to the frames by linguists. 

For frame identification, we use the PLMs encoder, to encode the context sentence $s_i$, starting with a \texttt{[CLS]} token and ending with \texttt{[SEP]}. Specially, we use the special token [T] to mark the span of the target:
\begin{equation}\label{e1}
	{I_i} = \texttt{[CLS]}c_0...c_m\texttt{[T]}t_1...t_{1+l}\texttt{[T]}c_{m+1}...c_{n-1}\texttt{[SEP]},
\end{equation}
where  $c_m$ denote $m\text{-}th$ token in $s_i$, $m \in [0,n)$, and the $l$ indicates the length of the target $t$.
Next, we obtain a vector representation for each token in $I_i$ and use the representation for the [CLS] token, \textit{i.e.}, ${C}$, as the sentence embedding representation since it is used as an aggregate representation for $I_i$:
\begin{equation}\label{e2}
	{h^0_i},{h^1_i},...{h^n_i} = encoder(I_i),
\end{equation}
\begin{equation}\label{i0}
	C = {h^0_i}.
\end{equation}
We obtain the position encoding matrix $\tilde{m_i}$ of the target and then compute the embedding representation of the target from $s_i$ by averaging the pooling of  $\tilde{m_i}$ and ${h^0_i}$.
The embedding representation of the target can be defined as
\begin{eqnarray}\label{e2}
	\tilde{m}_i&=&
	\left\{
	\begin{array}{lll}
		1 \  \ \hfill {c}_{i} \in \{t_1...t_{1+l}\} \\
		0 \  \ \hfill else \\
	\end{array}
	\right.
\end{eqnarray}
\begin{equation}\label{e3}
	\tilde{t_i} = f_{avg}({h^0_i} \cdot \tilde{m_i}),
\end{equation}
where $ f_{avg}(\cdot)$ is an averaging pooling function; $\tilde{t_i}$ is the embedding representation of the target.
Based on $\tilde{t_i}$, we feed it into the dense layers followed by a softmax layer to obtain a correctness $\hat{w_i}$ for each frame.
\begin{equation}\label{e4}
	w_i = tanh(\bm{W}\tilde{t_i}+ b),
\end{equation}
\begin{equation}\label{e5}
	\Omega = [\hat{w_0};...;\hat{w_{l-1}}] = \texttt{softmax}([{w_0};...;{w_{l-1}}])
\end{equation}
where $\bm{W}$ is a trainable matrix, $b$ is a trainable parameter. $l$ denotes the number of candidate frames.
During training, we minimize the cross entropy loss which measures the difference between $\Omega$ and the gold labels. During inference, we choose the frame in $\Omega$ with the highest score $\hat{w}$ as the correct semantic frame.

\subsubsection{Argument Identification}
Argument identification aims to identify the starting and ending positions of each argument $a_i$ in a sentence \cite{zheng2023query}. 
In this task, we consider it as a sequence labeling problem: given a sentence, identifying the start and end positions of frame elements concepts. 

Similarly to semantic frame modeling, we use language models to encode the sentence $s_i$. 
However, we innovate by directly using the embedded representation of the position [CLS], denoted as $h^0_i$, as input for the token classifier. When predicting tokens, we pass $h^0_i$ through a linear layer to predict the label with the highest probability. Formally, these operations can be defined as follows:
\begin{equation}\label{e6}
	a_i = tanh(\bm{W}h^0_i+ b),
\end{equation}
\begin{equation}\label{e7}
	\omega = [\hat{a_0};...;\hat{a_{n-1}}] = \texttt{argmax}([{a_0};...;{a_{n-1}}]),
\end{equation}
where $\omega_j$ is one of the “B-I-O” labels, $n$ denotes the length of tokens, $j \in [0,n)$.

\begin{wrapfigure}[23]{r}{0.55\textwidth}
\centering
\includegraphics[scale=0.4]{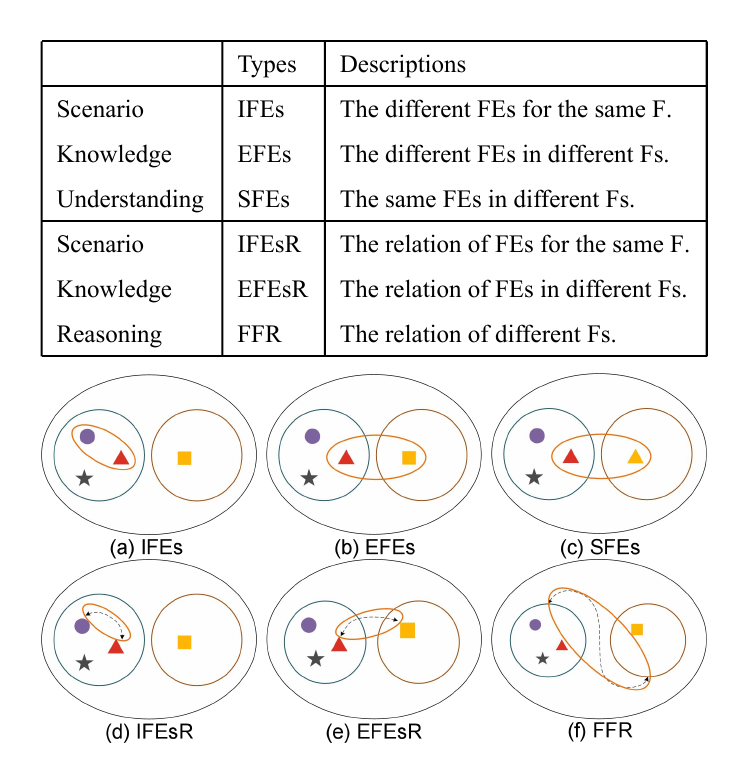}
\caption{An illustration of six different probing types. Note that they investigate the understanding of Internal FEs (IFEs), External FEs (EFEs), Same FEs (SFEs), and reasoning of Internal FEs (IFEsR), External FEs (EFEsR), and Frame-to-Frame (FFR). The two circles in the elliptical circle indicate two different frames, of which the $\star$, $\filledmedtriangleup$, and $\filledmedsquare$ indicate the different FEs of a frame.}
\label{figure4}
\end{wrapfigure}

\subsubsection{Frame Element Modeling}
We leverage the FEs in a frame to represent concepts in a scenario. 
Therefore, for frame element modeling, we aim to assign to a FE for the argument $a_i$, target $w_t$, and sentence $s$, where $a_i$ denote the $i$-th argument \cite{zheng2023query}.
We first utilize PLMs to obtain the embeddings of \texttt{[CLS]} as the sentence representation. Besides, we also use special tokens \texttt{[T]} and \texttt{[F]} to label the target and frame.
For the argument features, we obtain the token-level weights $w = \{w_j,w_{j+1},...,w_{j+c-1}\}$ of each argument from $h^i_0$, and then obtain the embedding representation of arguments $\hat{a}$ by
\begin{equation}\label{e8}
	r_{a} = f_{max}(\frac{1}{c}\sum_{i=j}^cw_i),
\end{equation}
where $ f_{max}(\cdot)$ is max pooling function; $c$ is the length of the argument; $w_j$ is the token-level weights. 
Finally, we feed $r_{a}$  into the MLP to predict the classification probability for each argument.
\begin{equation}\label{e9}
	p = \bm{w^T_2}relu(\bm{W_1}r_{a}+ \bm{b_1})+b_2,
\end{equation}
where $\bm{W_1}$ is a trainable matrix, $\bm{w_2}$ and $\bm{b_1}$ are trainable vectors, and $b_2$ is a trainable parameter.

\subsection{Frame-based Knowledge Graph}
To better estimate models using the knowledge beyond a text, we introduce a frame-based knowledge graph, denoted by $\mathcal{G}$.
It can be viewed as a subgraph composed of extracted frames and frame elements based on a given context. 

\noindent \textbf{Formalization. } The frame-based knowledge graph can be formalized as $\mathcal{G}=(\mathcal{V},\mathcal{E})$, where $\mathcal{V}$ is the node set and $\mathcal{E}$ is the edge set.
We utilize prior frame semantic parser to extract the semantic frames and frame elements and take them as the nodes of $\mathcal{G}$. Thus there are two types of nodes: $v^F$ and $v^{E}$, where $v^F \in \mathcal{V}$ and $v^{E} \in \mathcal{V}$. In particular, $v^F$ represents the frames. $v^{E}$ denotes the filling text related to frame elements in context.

For edges of $\mathcal{G}$, we consider three types of edges to explore the models' understanding of frames and semantic roles: $e^{F \text{-} F}$, $e^{F\text{-} E}$, and $e^{E\text{-} E}$, where
$e^{F \text{-} F}$ represents the relation between two frames. $e^{F\text{-} E}$ describes the relation between a frame and its frame elements, and $e^{E\text{-} E}$ indicates the relation between frame elements, which are from different frames or the same frames. 

\begin{figure}[t]
    \centering
    \includegraphics[scale=0.43]{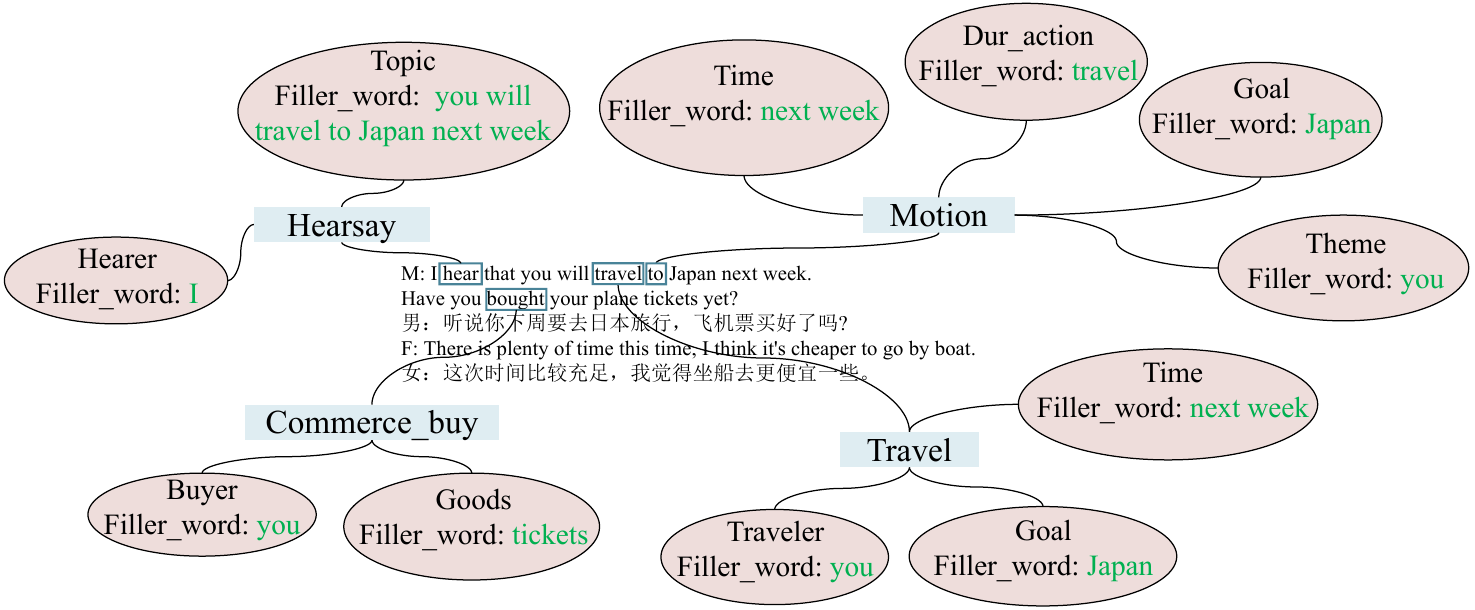} 
	\caption{An example frame-based knowledge graph on C3-D dataset \cite{c3}.}
    \label{figureframegraph}
\end{figure}

\noindent \textbf{Graph Construction.} 
As shown in Figure \ref{figureframegraph}, we provide an example of a frame-based knowledge graph constructed on a machine reading comprehension-based QA (MRC-QA) dataset \cite{c3}. 
The dataset comprises contexts and QA pairs. 
We first use the above frame semantic parser to extract the frames and semantic FEs involved in the context. Based on the graph formalization, we then construct a context subgraph. Notably, the semantic information represented in this graph is entirely derived from external semantic knowledge beyond the dataset itself. 

\subsection{Knowledge Probing}
Inspired by previous research \cite{petroni-etal-2019-language,wang-etal-2021-generative}, we convert the triple knowledge (e.g., \texttt{<you, Travel, Traveler>}) from the frame-based knowledge graph into the multiple-choice QA pairs (e.g., \texttt{In Travel scenario, you is a [mask]. A. Traveler. B. Reader. C. Hearer. Answer: A}) using prompt templates. We then utilize a questions-based way \cite{Kalo2022KAMELKA} to probe whether the black-box PLMs store implicit knowledge beyond the text. Finally, as shown in Figure \ref{figure4}, we consider probing six different hidden knowledge types, including the understanding of IFEs, EFEs, SFEs, and reasoning of IFEsR, EFEsR, and FFR. 

Specifically, given context $c$, question $q$ (or a set of candidate choices), and the answer $a$, we first generate frame-based hidden knowledge based on the above steps, including frame-based context $c^h$, question $q^h$, the frame-based answer $a^h$.
Subsequently, we employ a black-box PLM $f(\cdot)$ to assess QA performance by inputing $c$ and $q$, 
\begin{equation}\label{e10}
	\hat a = f(c;q).
\end{equation}
Next, we observe the QA performance on implicit knowledge by inputting $c^h$ and $q^h$ beyond the context $c$, 
\begin{equation}\label{e10}
	\hat a^h = f(c^h;q^h),
\end{equation} 
which is because we have a corollary empirically.
\begin{corollary}
For a machine reading comprehension-based QA task, given context $c$, question $q$ (or a set of candidate choices), and the answer $a$. if a model or human $f(\cdot)$ can always answer question $q$ based on context $c$ and give a prediction $\hat a = a$, then $f(\cdot)$ must understand context $c$ and question $q$. 
\end{corollary}

That is to say, $f(\cdot)$ stores unseen knowledge related context $c$ and question $q$.
Therefore, we prob this unseen knowledge to assess the performance of strong black-box PLMs.
We consider two scenarios. For the zero-shot (ZS) scenarios, we evaluate directly on two types of data. For fine-tuning (FT) scenarios, we first train on benchmark data, and then use the trained model weights to evaluate on this unseen knowledge beyond the text itself. 

\section{Experiments}
In the section, we explore and conduct experiments to address the four key issues:

\begin{itemize}
\setlength{\itemsep}{0pt}	
    \item \textbf{RQ1:} Do strong PLMs capture the hidden knowledge beyond a given text?
    \item \textbf{RQ2:} How can PLMs learn unseen knowledge beyond a given text based on the black-box embeddings?
    \item \textbf{RQ3:} Can frame-based knowledge modeled effectively enhance the representation learning abilities of PLMs? 
    \item \textbf{RQ4:} Do large-scale PLMs with strong emergent capabilities have inherent limitations for this hidden knowledge? 
\end{itemize}

\subsection{Experimental Models, Datasets and Metric}

\noindent \textbf{Experimental Models.} 
We investigate representative black-box models to explore, including small-scale and large-scale PLMs. For small language models (SLMs), we evaluate BERT \cite{2019bert}, XLNet \cite{xlnet}, MacBERT \cite{macbert}, ERNIE \cite{ERNIE3.0}. For large language models (LLMs), we perform ChatGPT (gpt-3.5-turbo) \cite{chatgpt}, ChatGLM (ChatGLM-6B) \cite{du2022glm}, ERNIE-based models (ERNIE-4.0-8K, ERNIE-3.5-128K), LLaMa-based models (LLaMa-2-7B, LLaMa-2-13B, LLaMa-3-8B) \cite{dubey2024llama,touvron2023llama}, Gemma-7B-It \cite{team2024gemma}, Mixtral-8x7B \cite{jiang2024mixtral}, and XuanYuan-70B \cite{zhang2023xuanyuan}. 

\noindent \textbf{Datasets and Metric.} 
We consider QA task as an example, including three machine reading comprehension-based multi-choice QA datasets \cite{c3}, a mixed-genre dataset (C3-M), dialogue-based dataset (C3-D), and the dataset they merged (C3-ALL). They provide given context that can be used to mine unseen frame-based knowledge. 
This enables us to validate the significance of frame-based knowledge modeling in evaluation across diverse scenarios. 
For evaluation metrics, we adopt accuracy (Acc) for the multi-choice QA task. 

\noindent \textbf{Implementation Details.} 
We consider the zero-shot and fine-tuning scenarios to implement the above models. For model training, the Adam \cite{P2015Adam} is used as an optimizer; the learning rate is initialized to 2e-5; the batch size is set to 8 or 20; the warmup proportion is 0.1; maximum sequence length is 256 or 310. We conduct all our experiments on NVIDIA V100 GPUs with 32G. 

\subsection{Experiments Results}

\begin{table}[h] \centering
\caption{Performance (\%) of LMs on the random and zero-shot learning on \texttt{KnowProb}. For the random scenario, the experiments are repeated multiple times (5 times) and averaged.}\label{tabzeroshot}
\begin{tabular}{l|l|cc|cc|cc}
\hline
        &  & \multicolumn{2}{c|}{C3-M Dataset} & \multicolumn{2}{c|}{C3-D Dataset} & \multicolumn{2}{c}{C3-ALL Dataset} \\

ID & Models  & Dev & Test &Dev & Test  & Dev & Test  \\
\hline
0 & Random  & \textbf{33.254} & \textbf{33.104} & \textbf{36.293} & \textbf{35.742} & \textbf{34.909} & \textbf{34.423} \\
\hline
  1 & BERT & 32.979 & 32.788 & 33.572 & 33.097 & 33.336 & 32.892 \\ 
	2 & XLNet& 25.006 & 24.350 & 23.893 & 23.887 & 24.334 & 24.715 \\ 
3 & MacBERT& 20.375 & 20.510 & 22.698 & 22.274 & 21.506 & 21.387 \\ 
4 & ERNIE& 22.052 & 22.352 & 22.712 & 22.921 & 22.205 & 22.418 \\ 
\hline
		& Avg  & 25.103	& 25.000& 	25.719& 	25.545& 	25.345& 	25.353 \\
\hline
\end{tabular}
\end{table}
\noindent  \textbf{RQ1: Exploring the ability of language models to capture the hidden knowledge behind a given text.} To investigate whether language models encode the hidden knowledge beyond the given text, we conduct experiments on representative models using generated knowledge by \texttt{KnowProb} under both random and zero-shot scenarios. For the random scenario, we randomly select one option from multiple choices as the predicted answer. For the zero-shot scenario, we directly utilize language models from HuggingFace\footnote{https://huggingface.co/} to answer the question $q^h$ related to hidden knowledge (e.g., \texttt{"Is Tom’s grandma the reader?"} in Figure \ref{figure1}), without performing any fine-tuning. 
Table \ref{tabzeroshot} illustrates the outcomes of our experiments, showcasing the results for both random and zero-shot scenarios. Upon inspection, it becomes evident that, in the absence of any downstream task fine-tuning, the performance of language models falls short of the random results. This observation points towards the limitation of language models in effectively capturing the hidden knowledge beyond a text itself. 

\begin{table}[h] \centering
\caption{Performance (\%) of fine-tuned models on hidden knowledge using C3 benchmark.}\label{tabft}
\begin{tabular}{l|l|cc|cc|cc}
\hline
        &  & \multicolumn{2}{c|}{C3-M Dataset} & \multicolumn{2}{c|}{C3-D Dataset} & \multicolumn{2}{c}{C3-ALL Dataset} \\

ID & Models  & Dev & Test &Dev & Test  & Dev & Test  \\
\hline

        1 & BERT & 38.335 &	37.876&	36.236&	36.152&	38.022&	37.923 \\ %
		2 & XLNet& 34.483 &	34.221&	38.728&	37.959&	38.352&	38.016 \\ %
		3 & MacBERT& 39.249	&38.683&	39.075&	38.987&	39.885&	39.702 \\ %
		4 & ERNIE& 38.414 &	38.155&	38.186&	37.868&	39.458	&39.142\\ %
\hline
			& Avg  & {37.620}  &	{37.234} &	{38.056}	 &{37.742}	 &{38.929} &	{38.696} \\
\hline
\end{tabular}
\end{table}

\begin{figure*}[t]
\centering
\subfloat[ZS using BERT]{
\includegraphics[width=0.24\textwidth]{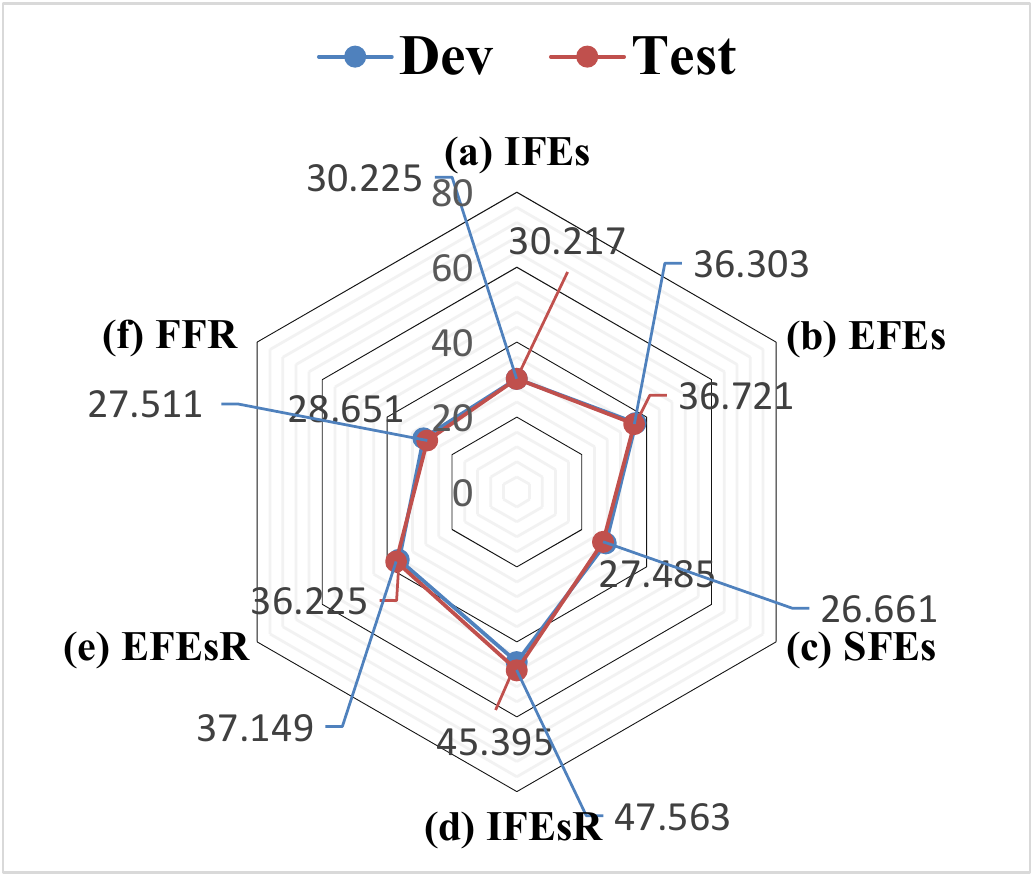}
}
\subfloat[ZS using XLNet]{
\includegraphics[width=0.24\textwidth]{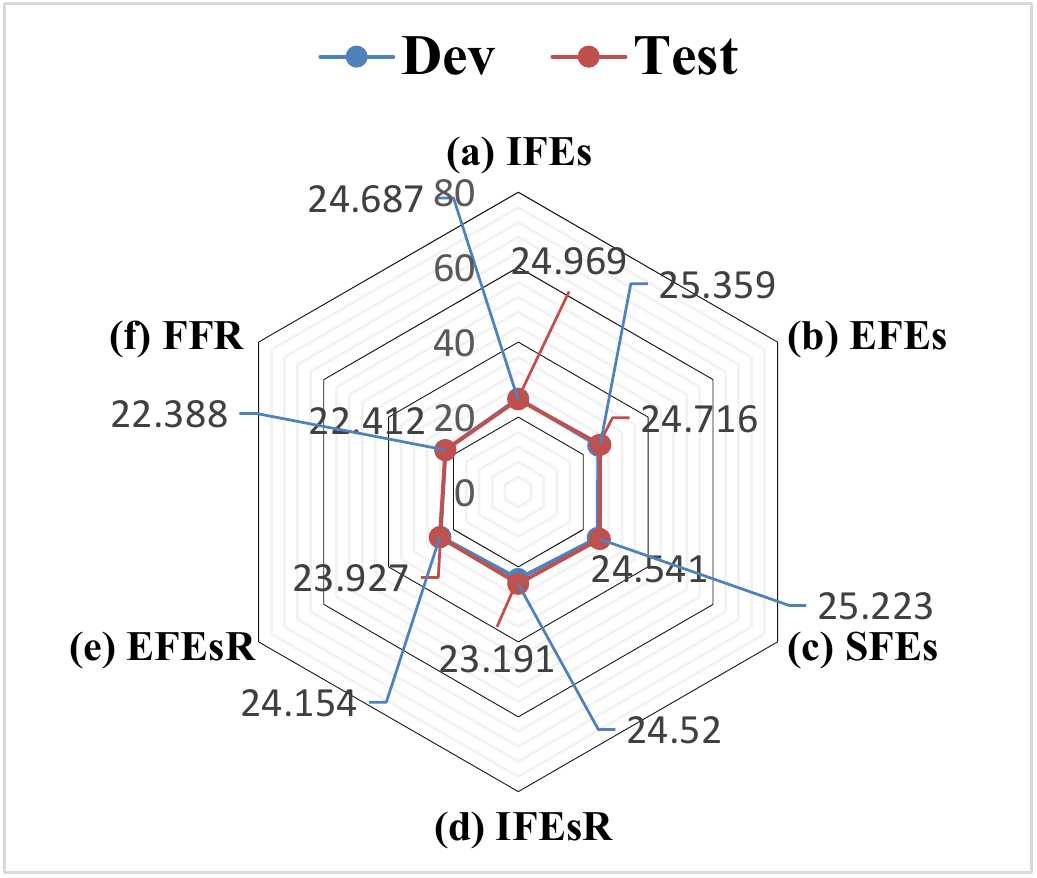}
}
\subfloat[ZS using MacBERT]{
\includegraphics[width=0.24\textwidth]{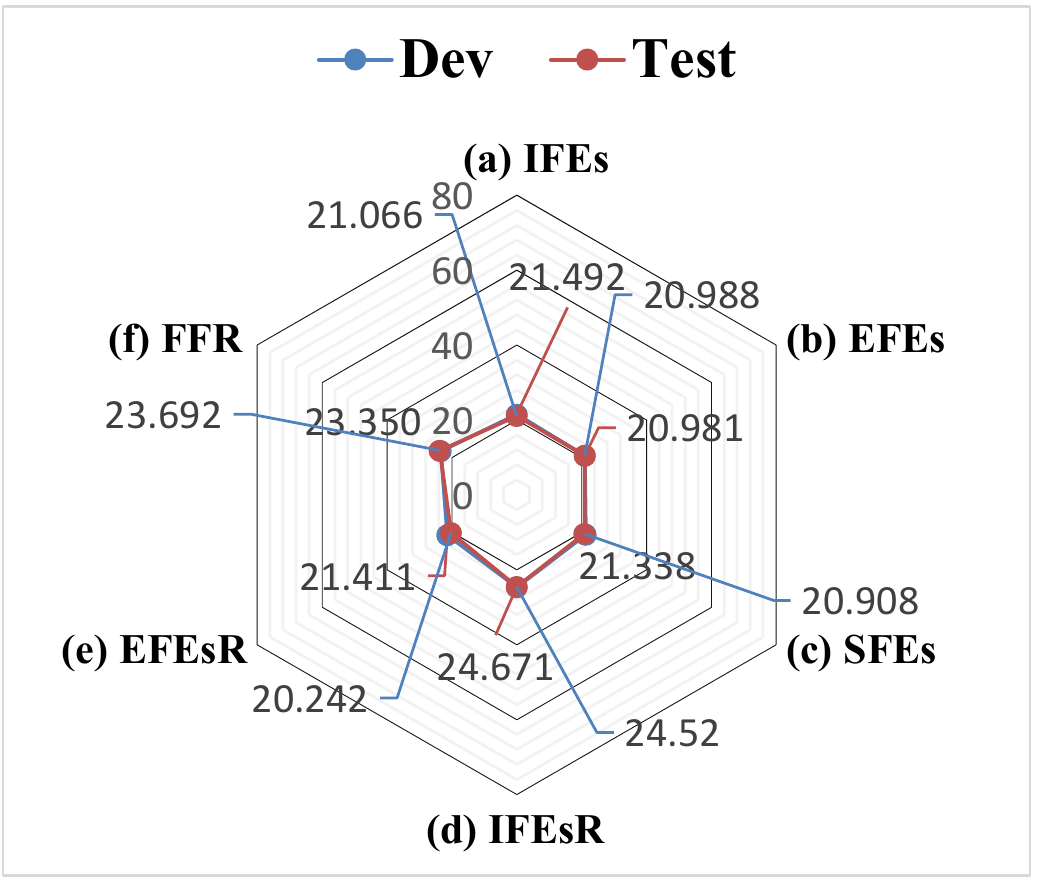}
}
\subfloat[ZS using ERNIE]{
\includegraphics[width=0.24\textwidth]{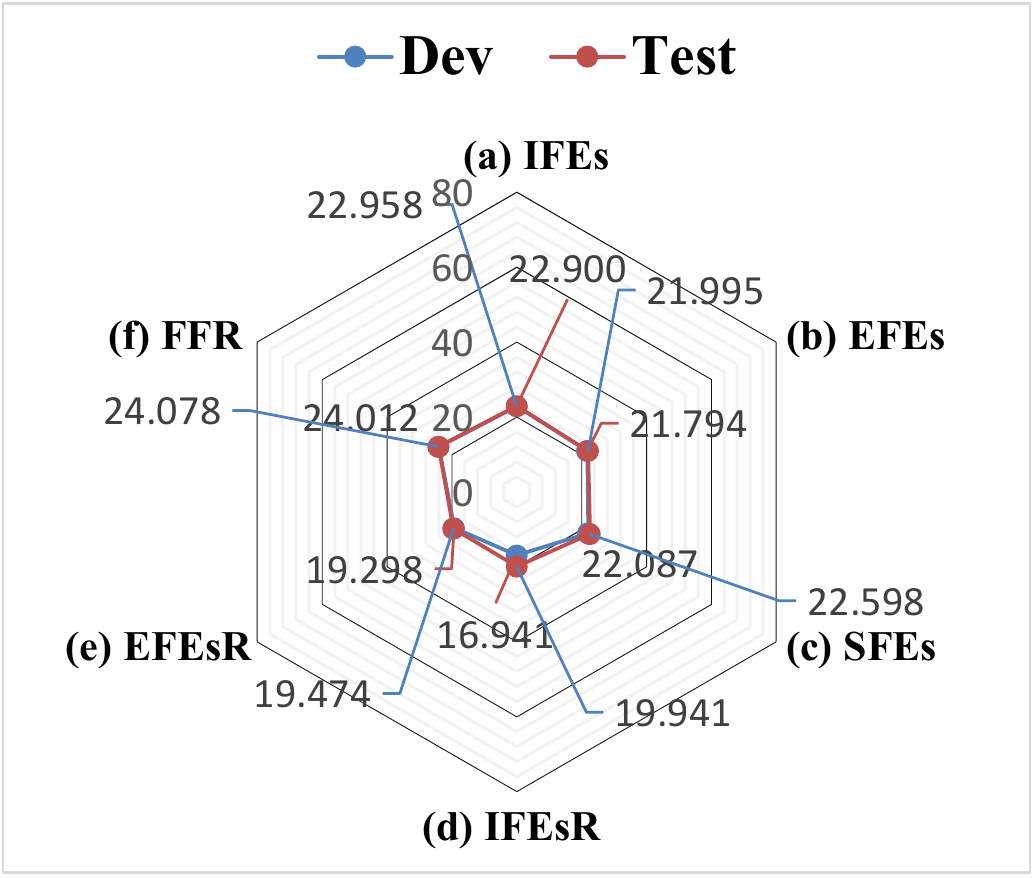}
}
\hspace{-9mm}
\subfloat[FT using BERT]{
\includegraphics[width=0.24\textwidth]{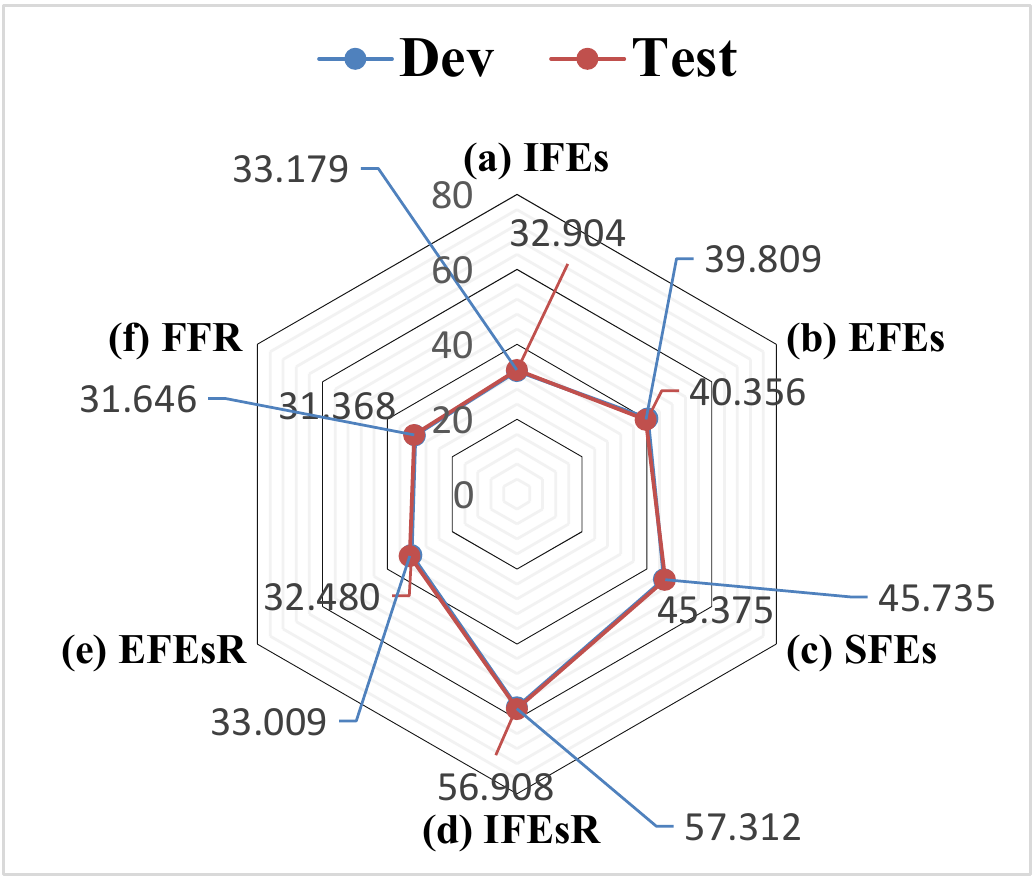}
}
\subfloat[FT using XLNet]{
\includegraphics[width=0.24\textwidth]{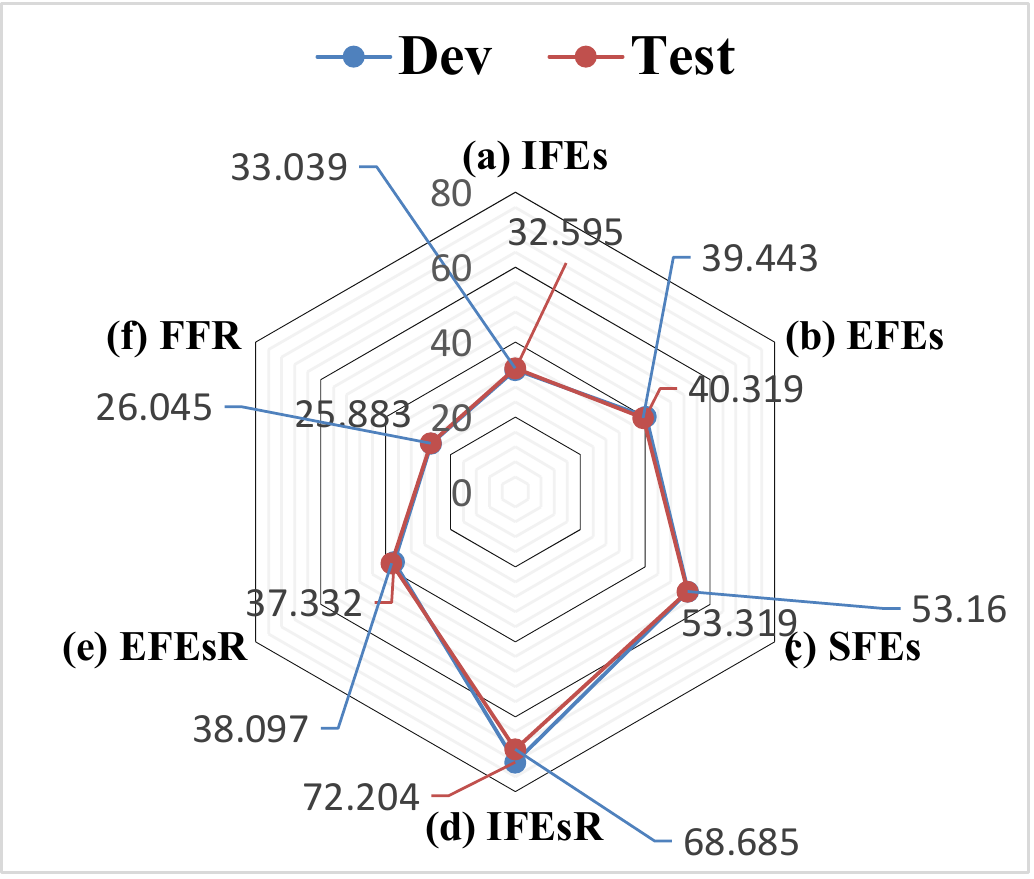}
}
\subfloat[FT using MacBERT]{
\includegraphics[width=0.24\textwidth]{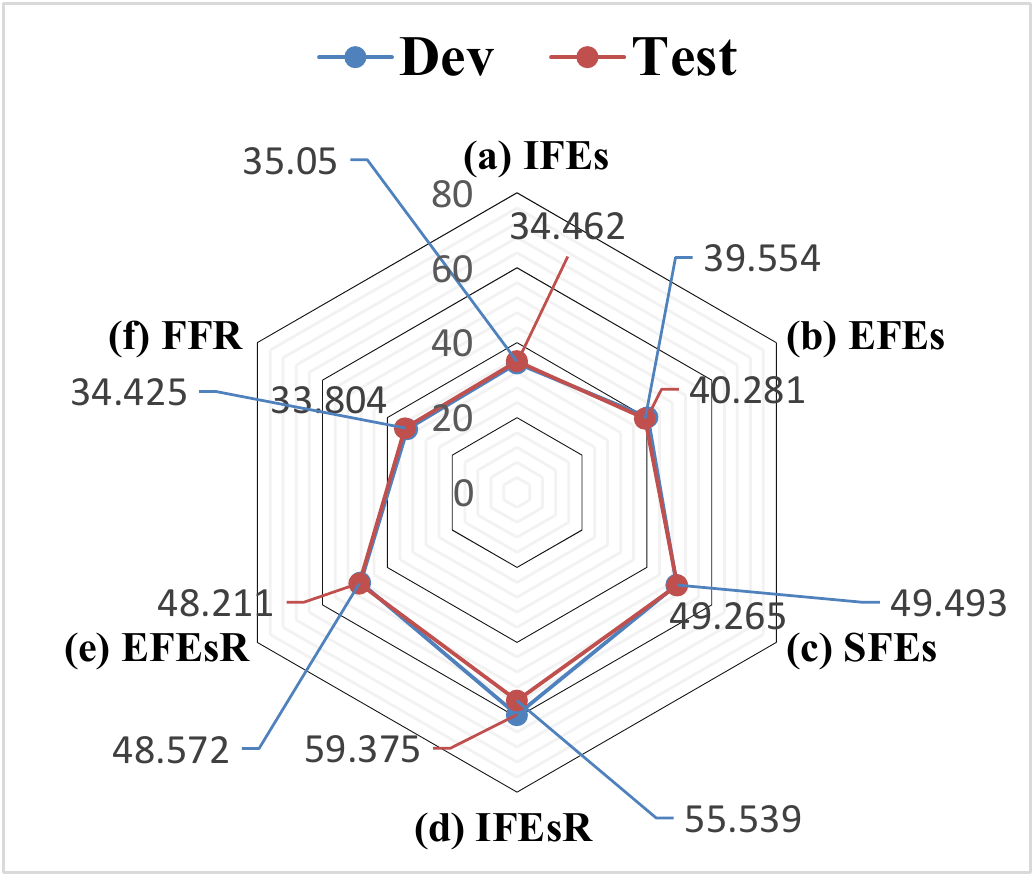}
}
\subfloat[FT using ERNIE]{
\includegraphics[width=0.24\textwidth]{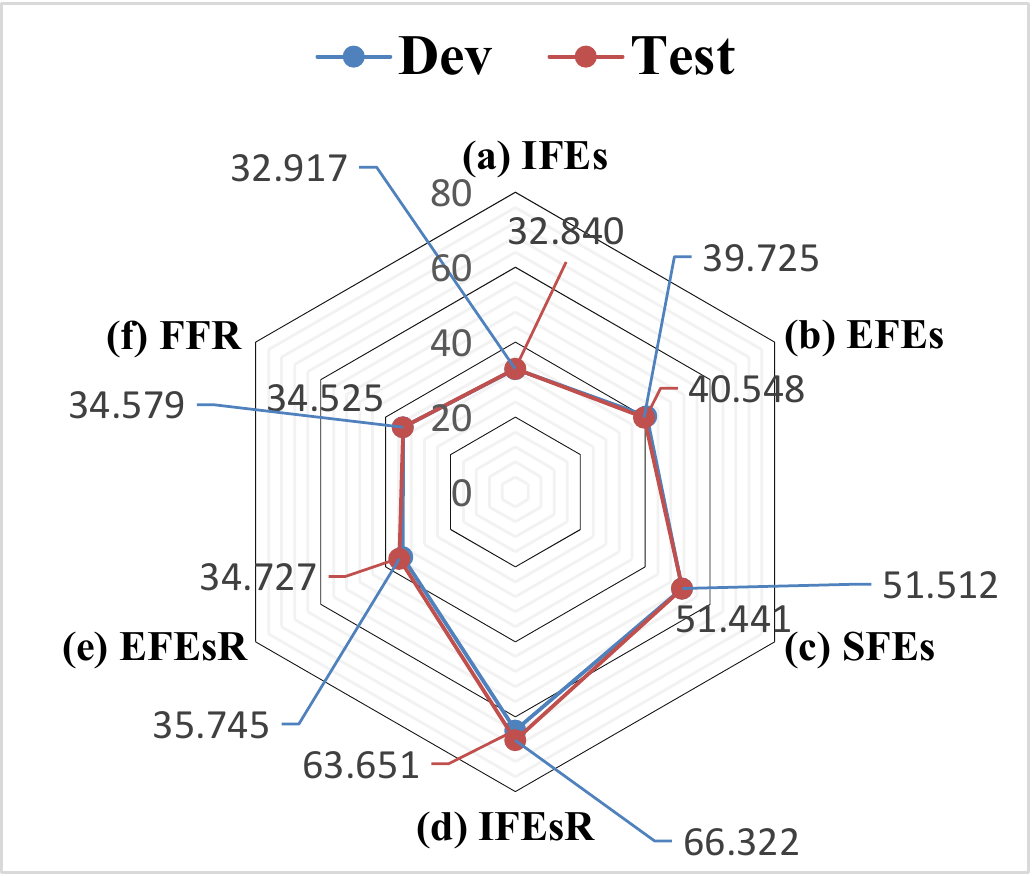}
}
\caption{Performance (\%) of different evaluation types using \texttt{KnowProb} on C3-ALL in detail. We use \texttt{KnowProb} to probe models in two scenarios: (a) ZS: LMs without any task-specific fine-tuned training; (b) FT: fine-tuning models using the given benchmark.}
\label{figrudar}
\end{figure*}
\noindent \textbf{RQ2: Fine-tuning helps PLMs learn unseen knowledge beyond a given text.} Firstly, we use the standard fine-tuning approach to train models on the original QA dataset and evaluate the fine-tuned models (i.e., BERT, XLNET, MacBERT, ERNIE) using \texttt{KnowProb}. 
Table 2 shows the experimental results from \texttt{KnowProb}.
Surprisingly, despite the absence of any frame-based hidden knowledge or sample generated by \texttt{KnowProb} in the training data, the fine-tuned models still exhibited significant improvements in capturing hidden knowledge (from 25.000/25.545/25.353 to 37.234/37.742/38.696 on C3-M/C3-D/C3-ALL). This suggests that PLMs when fine-tuned on surface-level text, can effectively conduct representation learning for some hidden knowledge beyond a given text. 
Furthermore, we delved into the limitations of the models on six dimensions. As depicted in Figure \ref{figrudar}, the probing results on multiple models across different types are shown. We observed that \texttt{KnowProb} can provide more detailed insights. Through experimental results, we found that fine-tuned language models resulted in remarkable improvements in various dimensions, with a more significant improvement in the IFEsR. This indicates that PLMs benefit significantly from both pre-training and fine-tuning paradigm, and the fine-tuning approach helps the model prioritize establishing connections between a word and the knowledge within the same hierarchy during representation learning.

\begin{figure*}[t]
\centering
\subfloat[CEFT using BERT]{
\includegraphics[width=0.24\textwidth]{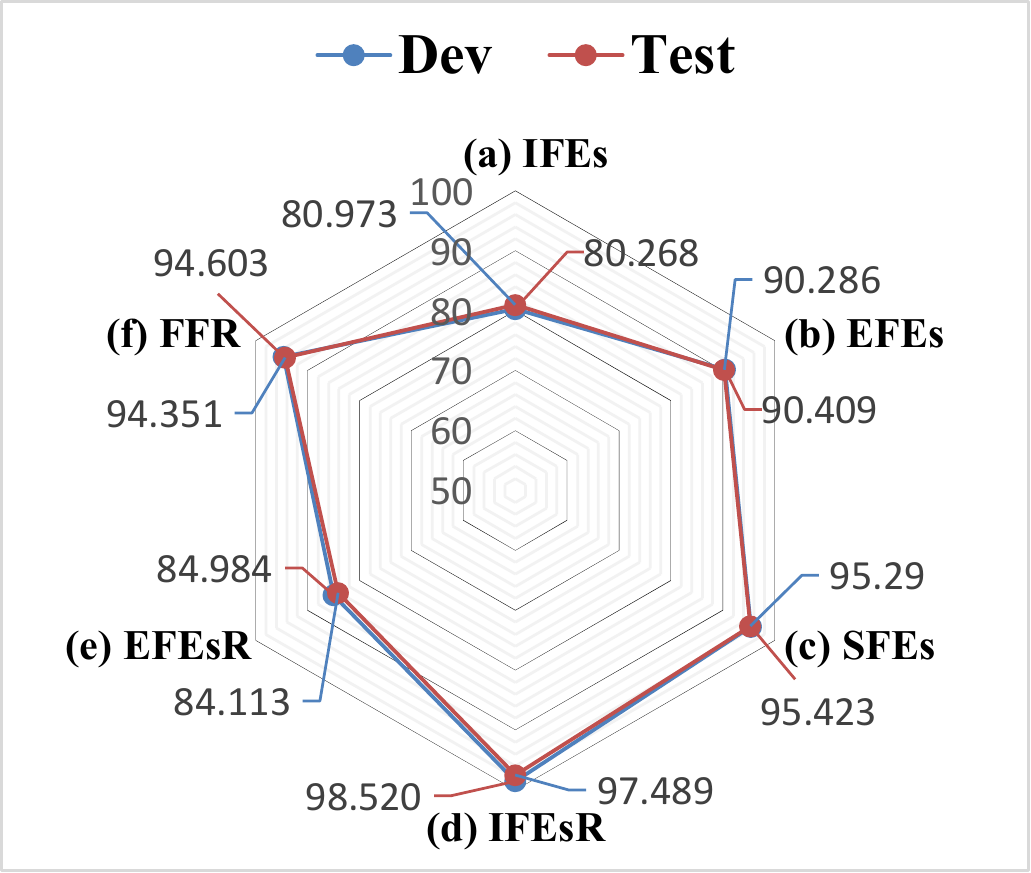}
}
\subfloat[CEFT using XLNet]{
\includegraphics[width=0.24\textwidth]{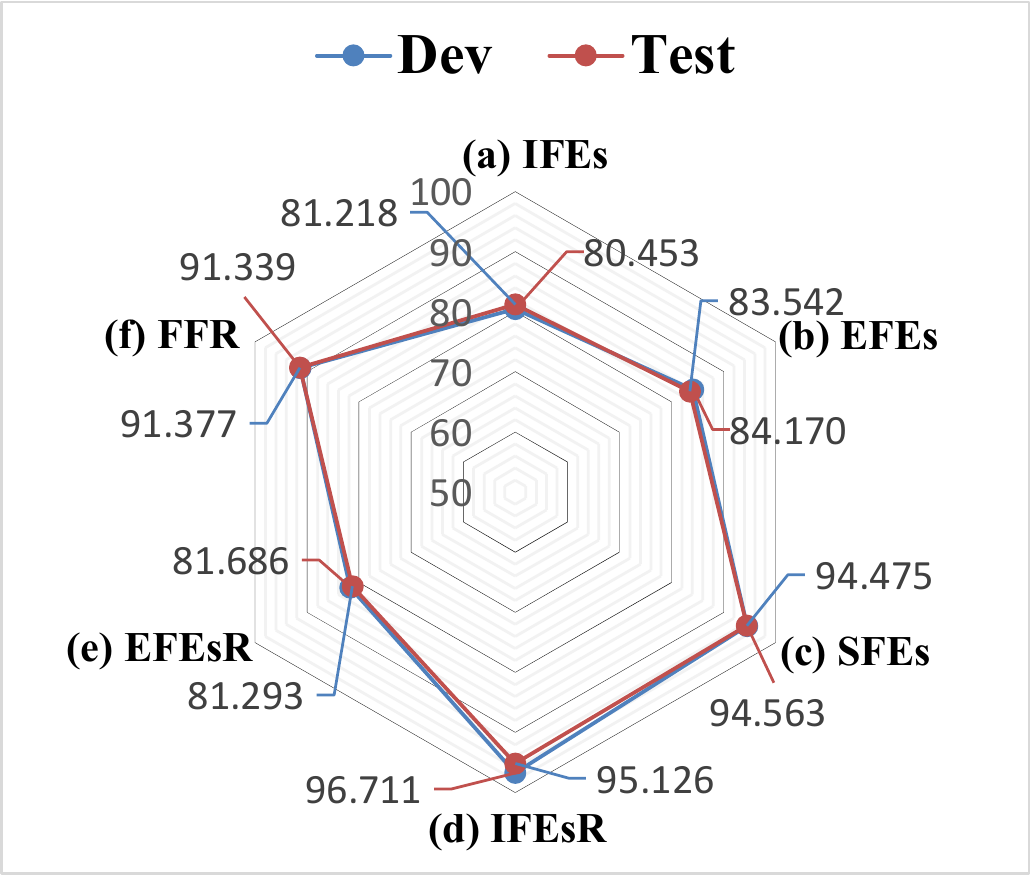}
}
\subfloat[CEFT using MacBERT]{
\includegraphics[width=0.24\textwidth]{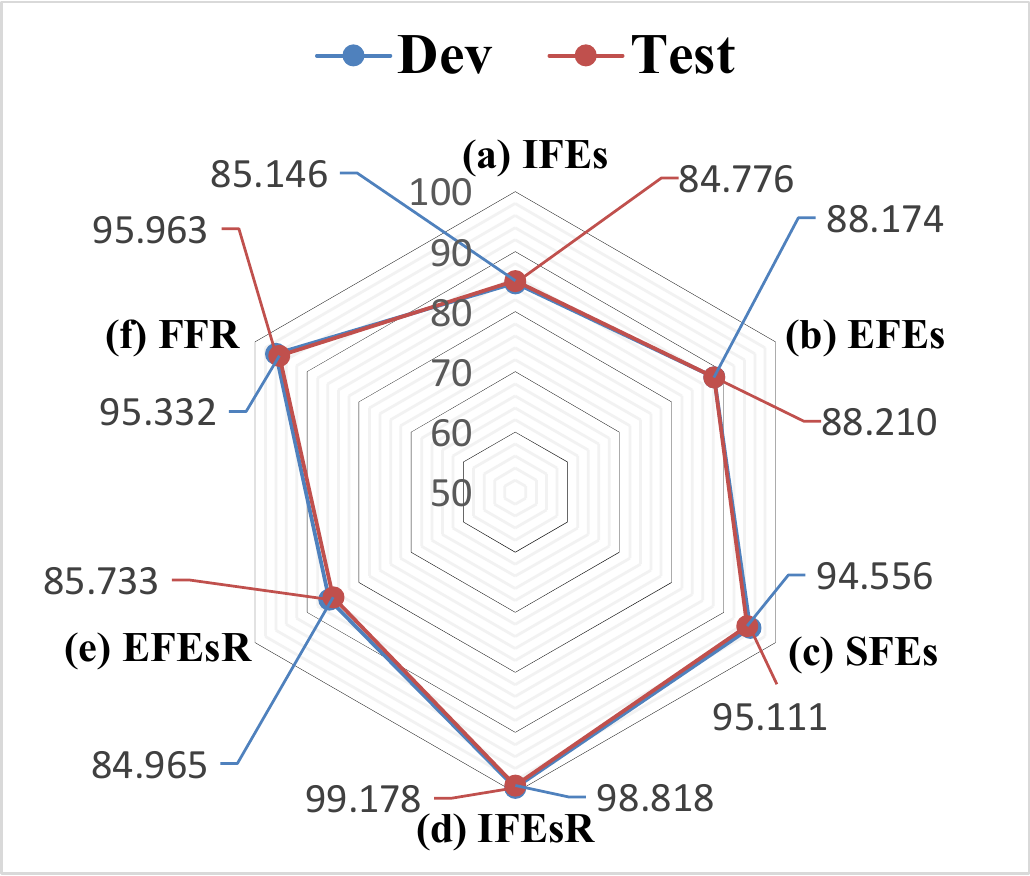}
}
\subfloat[CEFT using ERNIE]{
\includegraphics[width=0.24\textwidth]{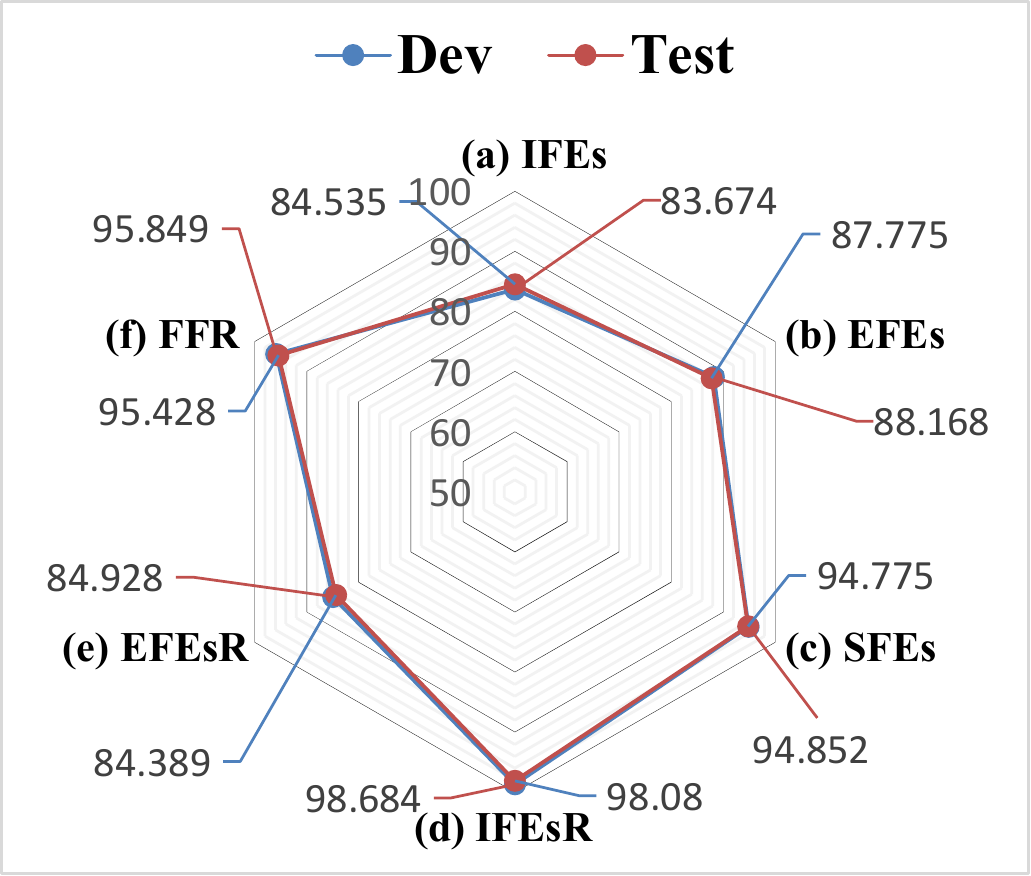}
}
\caption{Performance (\%) of different probing types on \texttt{KnowProb} using C3-ALL in detail. We use \texttt{KnowProb} to probe models in the scenarios: CEFT: fine-tuned models with hidden knowledge-enhanced training. }
\label{figrudar1}
\end{figure*}

\noindent \textbf{RQ3: Enhancing the representation learning of PLMs with frame-based knowledge.} 
To verify the effectiveness of the generated frame knowledge, we conduct an enhanced experiment. 
Figure \ref{figrudar1} shows the fine-tuned experimental results with frame knowledge-enhanced training. 
We can observe that the integration of frame-based knowledge present has shown significant improvements across multiple models for the representation of hidden knowledge. 
\begin{wraptable}[14]{r}{0.5\textwidth}
  \centering
  \caption{Performance (\%) of LLM using \texttt{KnowProb} on C3-M and C3-D datasets.}\label{tabLLM}
  \resizebox{0.40\textwidth}{!}{
  \begin{tabular}{l|c|c}
    \bottomrule
  Model & C3-M (\%) & C3-D (\%)\\
    \bottomrule
    \hline
    GPT-3.5-turbo & 38.6 & 40.2 \\
    \hline
    ChatGLM-6B & 40.6 & 43.4 \\
    \hline
    ERNIE-4.0-8K & \textbf{58.3} & \textbf{65.0} \\
    
    ERNIE-3.5-128K & 50.2 & 60.0 \\
    \hline
    LLaMa-2-7B & 40.2 & 41.8 \\
    
    LLaMa-2-13B & 39.4 & 35.2 \\
    
    LLaMa-3-8B & 38.6 & 38.4 \\
    \hline
    Gemma-7B-lt & 32.4 & 38.8 \\
    \hline
    Mixtral-8x7B & 50.4 & 47.2 \\
    \hline
    XuanYuan-70B & 55.8 & 56.4 \\ 
   
    \bottomrule
\end{tabular}}
 \label{variants}
\end{wraptable}
In addition, as shown in Table \ref{tabablation}, we find that the surface-level QA performance has also achieved competitive results. 
Note that we do not employ complex reasoning logic setups; instead, we simply 
feed the generated knowledge to the model in a data augmentation manner. 
In particular, during the model training phase, the augmented data space is derived solely from hidden knowledge generated from the training set. 
This further suggests that the frame-based hidden knowledge generated is valuable and it has enormous potential for enhancing the representation learning capabilities of the models.

\begin{table}[t] \centering
\caption{The enhanced experiment (\%) on C3 Benchmark. The unseen knowledge indicates the hidden knowledge beyond the text itself.}\label{tabablation}
\begin{tabular}{l|l|cc|cc|cc}
\hline
        &  & \multicolumn{2}{c|}{C3-M Dataset} & \multicolumn{2}{c|}{C3-D Dataset} & \multicolumn{2}{c}{C3-ALL Dataset} \\

ID & Models  & Dev & Test &Dev & Test  & Dev & Test  \\
\hline

       1 & BERT & 58.463 & 57.043 & 61.973 & 62.857 & 62.683 & 63.591 \\ 
         &  + \textit{unseen knowledge} & {59.367} & {59.191} & {62.740} & {64.074} & {63.181} & {63.618} \\ 
\hline
		 2 & XLNet & 56.052 &	58.241	& 57.534&	56.085	&61.242	&61.562 \\ 
 &  + \textit{unseen knowledge} & {59.518}	&{58.492}	&{63.890}	&{61.958}	&{63.574}	&{62.050} \\ 
\hline
    3 & MacBERT & 62.682&	{62.987}	&64.219	&{63.492}	&65.303&	{66.007} \\ 
     &  + \textit{unseen knowledge} & {63.284}	&61.838	&{64.603}	&62.275&	{67.165}&	65.416 \\ 
\hline
     4 & ERNIE & 70.266	&71.678	&73.150&	73.122&	74.056&	73.869 \\ 
      &  + \textit{unseen knowledge} & {70.668}	&{72.328}	&{76.438}	&{73.968}	&{75.891}&	{74.615} \\ 
\hline
\end{tabular}
\end{table}

\noindent \textbf{RQ4: Probing hidden knowledge beyond the text itself for LLMs and Human.}
For the evaluation of humans, we randomly sample from the frame-based data 15 times to obtain 15 different subsets, and we invite 15 graduate students. In this process, these participants are provided with questions, choices, and context. Finally, the average accuracy of human is 92.418\%, and the upper bound for human performance is 96.552\%.

\begin{figure}[t]
    \centering
    \includegraphics[scale=0.365]{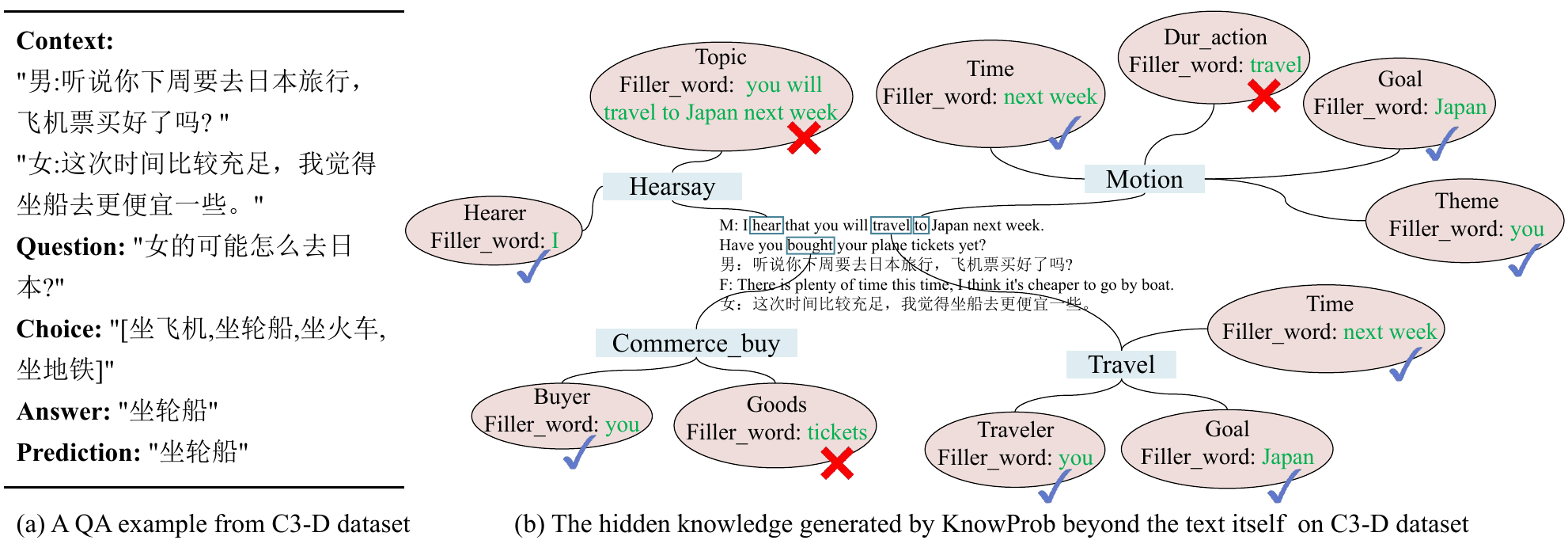}  
	\caption{A case example to prob LLaMa-2-13B on C3-D dataset (A Chinese dataset), including surface level QA data and generated hidden knowledge by \texttt{KnowProb}. Note that these terms and explanations. Motion: 位移. Travel: 旅行. Hearsay: 传闻. Commerce\_buy: 商业购买. Theme(转移体)/Traveler(旅行者): the entity undergoing location change. Goal(终点): the endpoint of the motion or travel. Time(时间): the time at which the motion or travel occurs. Dur\_action(持续时间): duration of motion's persistence. Buyer(买方): an individual who requires goods and exchanges money with the Seller for them. Goods(商品): anything that can be exchanged for money (including labor and time). Hearer(听者): the person who receives a Message. Message(信息): the content that is communicated from one person to another. Topic(主题)：the subject matter about which the Message is communicated. 
    }
    \label{figurecase}
\end{figure}

For LLMs probing, we estimate the performance of \texttt{GPT-3.5-turbo}, ChatGLM-6B, ERNIE-based models, LLaMa-based models, \texttt{Gemma-7B-It}, \texttt{Mixtral-8x7B} and \texttt{XuanYuan-70B} using \texttt{KnowProb} on test dataset of C3 benchmark. Due to the resource limitations of LLMs, we randomly sample 5 times from previous experimental data for evaluation. 
As shown in Table \ref{variants}, LLMs have a certain ability to capture hidden knowledge, while their power is not strong enough. Compared to the previous models, ERNIE-4.0-8K achieved the performance of the state of the art, which may be facilitated by its huge amount of parameters.
The results indicate, once again, that LLMs exhibit a certain level of hidden knowledge capture capabilities, but their abilities remain limited. The above analysis highlights \texttt{KnowProb} is effective for identifying the limitations of existing black-box models in an explainable way. 
Furthermore, we also provide some visual examples to showcase our proposed method. 
As shown in Figure \ref{figurecase}, given the context: \texttt{"男:听说你下周要去日本旅行，飞机票买好了吗?女:这次时间比较充足，我觉得坐船去更便宜一些。"}, we observe that LLaMa-2-13B can easily and correctly answer the question and understand that the woman may consider travelling to Japan by boat. However, it faces limitations when it comes to knowledge that goes beyond the text itself, such as \texttt{Tickets are Goods; Motion to Japan aims to Travel (an action).}
Although language models learn a distributed representation, this highlights opportunities for improving reasoning across out-of-domain distributions. 

\section{Conclusions}
In this paper, we propose \texttt{KnowProb}, a knowledge probing with frame-based knowledge modeling. We focus on investigating whether black-box PLMs understand hidden knowledge beyond
the text itself and provide six potential explanations derived from the underlying content of the given text. 
Extensive experimental results show that current small-scale (or large-scale) PLMs only learn a single distribution of representation and still face significant challenges in capturing the hidden knowledge behind a given text. 
Furthermore, we demonstrate that our proposed approach is effective in identifying the limitations of existing black-box models in a manner that is both explainable and rooted in cognitive linguistic principles. This facilitates community researchers to promote knowing black-box models and uncover black-box models. 

\section*{Acknowledgements}
We thank all anonymous reviewers for their valuable feedback. 
This work was supported by the National Natural Science Foundation of China (Nos.62376144, 61906111), the Science and Technology Cooperation and Exchange Special Project of Shanxi Province (No.202204041101016) and the Key Research and Development Project of Shanxi Province under Grant (No.202102020101008).

\bibliography{reference}
\bibliographystyle{splncs04}

\end{CJK*}
\end{document}